\documentclass[runningheads]{llncs}
\usepackage{graphicx}
\usepackage{amssymb}
\usepackage{amsmath}
\usepackage{stmaryrd}
\usepackage{multirow}
\usepackage{tabularx}
\usepackage{booktabs}
\usepackage{bbm}

\DeclareMathOperator*{\argmin}{arg\,min}
\begin{document}
\title{Inter Extreme Points Geodesics for End-to-End Weakly Supervised Image Segmentation}
\titlerunning{Inter Extreme Points Geodesics for Weakly Supervised Segmentation}
%

 \author{Reuben Dorent, Samuel Joutard, Jonathan Shapey, Aaron Kujawa, Marc Modat, S\'ebastien Ourselin and Tom Vercauteren}

\authorrunning{R. Dorent et al.}
%
 \institute{School of Biomedical Engineering and Imaging Sciences, King’s College London, London, United Kingdom
\email{reuben.dorent@kcl.ac.uk} 
}
\maketitle              
\begin{abstract}
We introduce \textit{InExtremIS}, a 
weakly supervised 3D approach to
train a
deep image segmentation
network
using particularly weak train-time annotations:
only 6 extreme clicks at the boundary of the objects of interest.
Our fully-automatic method is trained end-to-end and
does not require any test-time annotations.
From the extreme points, 3D bounding boxes are extracted around objects of interest. Then, deep geodesics connecting extreme points are generated to increase the amount of ``annotated" voxels within the bounding boxes. Finally, a weakly supervised regularised loss derived from a Conditional Random Field formulation is used to encourage prediction consistency over homogeneous regions. Extensive experiments are performed on a large
open
dataset for Vestibular Schwannoma segmentation. \textit{InExtremIS} obtained competitive performance, approaching full supervision and outperforming significantly other weakly supervised techniques based on bounding boxes. Moreover, given a fixed annotation time budget, \textit{InExtremIS} outperformed full supervision. Our code and data are available online.
\end{abstract}

\section{Introduction}
Convolutional Neural Networks (CNNs) have achieved state-of-the-art performance for many medical segmentation tasks when trained in a fully-supervised manner, i.e. by using pixel-wise annotations. In practice, CNNs require large annotated datasets to be able to generalise to different acquisition protocols and to cover variability in the data (e.g. size and location of a pathology). Given the time and expertise required to carefully annotate medical data, the data-labelling process is a key bottleneck for the development of automatic image medical 
segmentation tools. To address this issue, growing efforts have been made to exploit weak annotations (e.g. scribbles, bounding boxes, extreme points) instead of time-consuming pixel-wise annotations.

Weak annotations have been used in different contexts. The first range of approaches uses weak annotations
at test-time
to refine or guide network predictions. These techniques are commonly referred to as interactive segmentation \cite{Gulshan:star,Rother:grabcut,Wang:bifseg,Wang:deepigeos,Mcgrath:semi,Xu:Deep} or semi-automated tools \cite{Khan:Extreme,Maninis:dextr,Wang:levelset}. Networks are typically trained with fully annotated images and weak annotations are only used at inference time.
In contrast, weakly supervised techniques are only trained with incomplete but easy-to-obtain annotations and operate in fully-automatic fashion at inference stage \cite{Kervadec:constraint,Kervadec:boundingbox,Rajchl:deepcut,Zhang:lung,Roth:dextr3d,Dorent:scribbleda,Lin:scribblesup}. Bounding boxes are the most commonly used weak annotations \cite{Kervadec:constraint,Kervadec:boundingbox,Rajchl:deepcut}. However, extreme points are more time-efficient while providing extra information \cite{Papadopoulos:extreme}. To our knowledge, only one extreme points supervision technique has been proposed for 3D medical image segmentation \cite{Roth:dextr3d}. This method alternates between pseudo-mask generation using Random Walker and network training to mimic these masks. This computationally expensive approach relies on generated pseudo-masks which may be inaccurate in practice.

In this work, we propose a novel weakly supervised approach to learn automatic image segmentation using extreme points as weak annotations during training, here a set of manual extreme points along each dimension of a 3D image (6 in total). The proposed approach is end-to-end and trainable via standard optimisation
such as stochastic gradient descent (SGD). The contributions of this work are three-fold. Firstly, extreme points along each dimension are automatically connected to increase the amount of foreground voxels used for supervision. To prompt the voxels along the path to be within the object, we propose to use the network probability predictions to guide the (deep) geodesic generation at each training iteration. Secondly, we employ a CRF regularised loss to encourage spatial and intensity consistency. Finally, extensive experiments over a publicly available dataset for Vestibular Schwannoma segmentation demonstrate the effectiveness of our approach. With only 6 points per training image, our weakly supervised framework outperformed other weakly supervised techniques using bounding boxes and approached fully supervised performance. Our method even outperformed full supervision given a fixed annotation time budget.

\begin{figure}[tb!]
\begin{center}
\includegraphics[width=\linewidth]{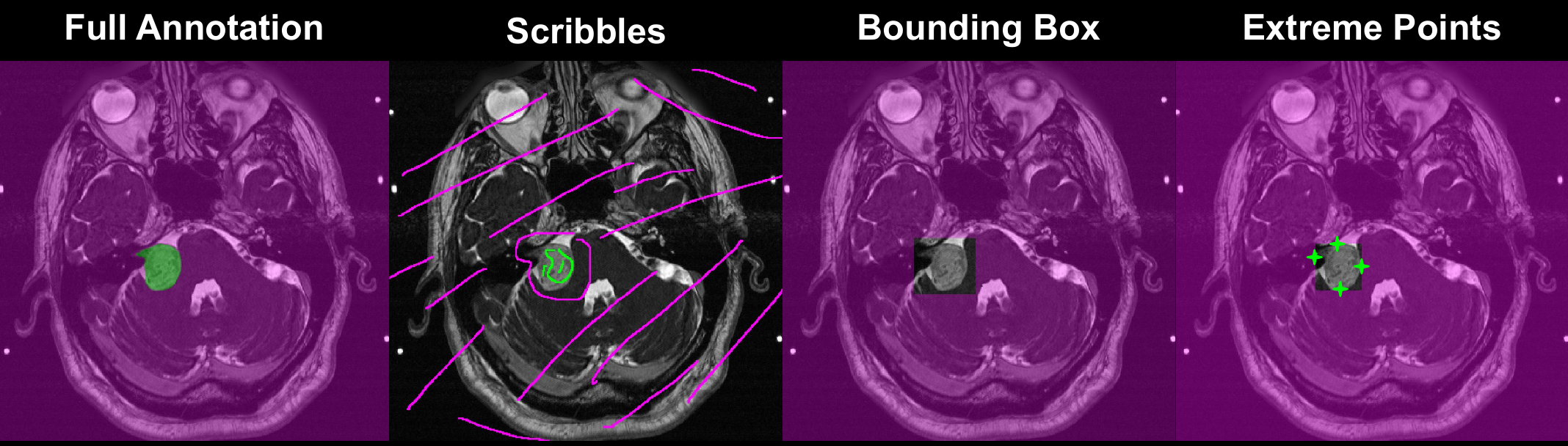}
\caption{Example of weak labels for our use case of Vestibular Schwannoma (VS) segmentation. For better readability, annotations are illustrated on a 2D axial slice. Magenta: Background. Green: VS.}
\label{example_annotations}
\end{center}
\end{figure}

\section{Related work}
\subsubsection{Weakly supervised image segmentation}
Weakly supervised learning covers a large variety of techniques that aim to build predictive models using time-efficient weak annotations. Most existing weakly supervised approaches \cite{Rajchl:deepcut,Zhang:lung,Roth:dextr3d} adopt an iterative training strategy that alternates between pseudo-mask (proposal) generations using non-learning techniques (e.g. Random Walker, CRFs) and training a CNN to mimic these proposals. The pseudo-masks are initially generated using the weak annotations and then refined using the CNN predictions. These techniques are computationally expensive and rely on robust but not always accurate proposal generation. For this reason, other studies investigated end-to-end approaches. Direct regularisation loss derived from CRFs has been proposed \cite{Tang:regularised,Dorent:scribbleda}, reaching almost full-supervision performance. This regularisation loss has only been employed with scribble annotations. Alternatively, constraints based on anatomical priors such as the volume or the shape have been studied for bounding box annotations \cite{Kervadec:boundingbox,Kervadec:constraint}. However, these approaches have only been tested on 2D slices with a bounding box provided per slice.

\subsubsection{Extreme points as weak annotations} Different weak annotations have been proposed such as scribbles \cite{Dorent:scribbleda,Lin:scribblesup,Tang:regularised,Wang:bifseg}, bounding boxes \cite{Rajchl:deepcut,Kervadec:boundingbox} and extreme points \cite{Maninis:dextr,Roth:dextr3d}. Examples are shown in Figure~\ref{example_annotations}. Scribbles are user-friendly and easy to use. However, they must not only cover the foreground but also the background. Since medical images are typically mostly composed of background, most of the annotation time is dedicated to the background. In contrast, bounding boxes and extreme points are focused on the object of interest.  Extreme clicks are the most time-efficient technique to generate bounding boxes and provide extra information compared to bounding boxes \cite{Papadopoulos:extreme}. A recent work proposed a weakly supervised technique using extreme clicks \cite{Roth:dextr3d}. This approach follows a computationally expensive two-step optimization strategy using Random Walker. To increase the foreground seeds, paths between the extreme points are initially computed, using only the image gradient information. In practice, this may lead to paths outside of the object and thus inaccurate initial pseudo-masks.

\section{Learning from extreme points}
\subsection{Notations}
Let $X: \Omega \subset \mathbb{R}^{3} \rightarrow \mathbb{R}$ and $Y: \Omega  \subset \mathbb{R}^{3} \rightarrow \{0,1\}$ denote a training image and its corresponding ground-truth segmentation, where $\Omega$ is the spatial domain. Let $f_{\theta}$ be a CNN parametrised by the weights $\theta$ that predicts the foreground probability of each voxel.
In a fully supervised learning setting, the parameters $\theta$ are optimized using a fully annotated training set $D=\{(X_i,Y_i)\}_{i=1}^{n}$ of size $n$.
However, in our scenario the segmentations $Y$ are missing.
Instead, a set of 6 user clicks on the extreme points $\mathcal{E}_{\text{points}}$ is provided for each training image $X$. 
These extreme points correspond to the left-, right-, anterior-, posterior-, inferior-, and superior-most parts of the object of interest.

Accurate extreme points further provide a straightforward means of computing
a tight bounding box $B_{\text{tight}}$ around the object. While voxels outside of the bounding box are expected to be part of the background, 
user-click
errors could result in some true foreground voxels lying outside the resulting bounding box. For this reason, a larger bounding box $B_{\text{relax}}$ is used for the background by relaxing the tight bounding box $B_{\text{tight}}$ by a fixed margin $r$. We hereafter denote $\overline{B_{\text{relax}}}$ the area outside the relaxed bounding box $B_{\text{relax}}$.

A straightforward training approach is to minimize the cross-entropy on the (explicitly or implicitly) annotated voxels, i.e. the 6 extreme points and 
$\overline{B_{\text{relax}}}$:
\begin{equation}
    \theta^* = \argmin_{\theta} \sum_{i=1}^{n} \bigg[ - \sum_{k \in \overline{B_{\text{relax}}}} \log\left(1-f_{\theta}\left(X_i\right)_{k}\right) - \sum_{k \in \mathcal{E}_{\text{points}}} \log\left(f_{\theta}\left(X_i\right)_{k}\right) \bigg] 
\end{equation}
This naive approach raises two issues: 1/ the problem is highly imbalanced with only 6 foreground annotated voxels;
and 2/ 
no attempt is made to propagate labels
to non-annotated voxels. The next two sections propose solutions to address these 
by using inter extreme points geodesics and a CRF regularised loss.
\begin{figure}[tb!]
\begin{center}
\includegraphics[width=\linewidth]{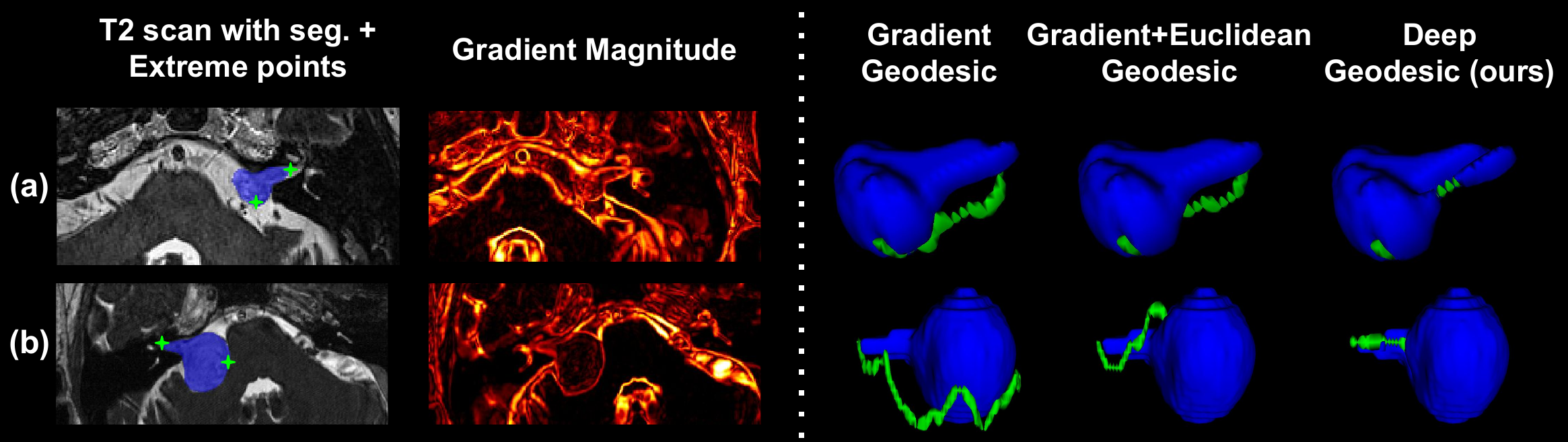}
\caption{Comparison of the different inter extreme points geodesics (green). Axial T2 slices with manual segmentation (blue) and image gradient magnitude are shown. (a) anterior- to posterior-most points geodesics; (b) right- to left-most points geodesics. Geodesics have been dilated for a better representation. Unlike the two other approaches, deep geodesics (ours) stay within the object. }
\label{example_geodesic}
\end{center}
\end{figure}
\subsection{Inter extreme points geodesics generation using CNNs output}
In this section, we propose to increase the amount of foreground voxels used for supervision by making better use of the extreme points. Let us assume that the foreground object is path-connected. Then, there is a path within the object that connects the extreme points. Voxels along this path could then be used for foreground supervision. At any given time during training, we take advantage of this connectivity hypothesis to compute inter extreme points paths which are 
likely
to remain in the foreground class
and use these for added supervision.

A popular approach to find paths connecting points within an object is based on geodesics or shortest paths with respect to a length metric. A geodesic is a length-minimising curve. In our case, the path domain is restricted to the tight bounding box $B_{\text{tight}}$. The length of paths intercepting edges in the image is typically penalised by accumulating the image gradient magnitudes \cite{Roth:dextr3d,Wang:deepigeos}. However, extreme points are on object boundaries and thus likely to be on intensity edges. The shortest path could then circumvent the object without intercepting any additional edge, as illustrated in Figure~\ref{example_geodesic}. Adding the Euclidean distance in the length metric helps \cite{Gulshan:star} but may still be insufficient.

To help geodesics remain in the object, we propose to penalise paths passing through background voxels using the network background probabilities. The length of a discrete path $\Gamma$ used to compute our deep geodesics is defined as:
\begin{equation}
      L_{\text{deep}}(\Gamma,X)=\sum_{k=1}^{|\Gamma|-1} \underbrace{\gamma_{e}d\left(\Gamma_{k}, \Gamma_{k+1}\right)}_{\text{Euclidean}}+\underbrace{\gamma_{g}\| \nabla X\left(\Gamma_{k}\right) \|}_{\text{Image Gradient}}+\underbrace{(1-f_{\theta}(X)_{k})}_{\text{CNN Background Prob.}}
\end{equation}
where $d\left(\Gamma_{k}, \Gamma_{k+1}\right)$ is  the Euclidean distance between successive voxels, and $\| \nabla X\left(\Gamma_{k}\right) \|$ is a finite difference approximation of the image gradient between the points $(\Gamma_{k},\Gamma_{k+1})$, respectively weighted by $\gamma_{e}$ and $\gamma_{g}$.

We additionally propose an automated hyper-parameter policy for $\gamma_{e}$ and $\gamma_{g}$ by scaling each quantity between $0$ and $1$, i.e.: 
\begin{equation*}
    1/\gamma_e=\max_{k\in B_{\text{tight}}} d(\Gamma_0, k) \text{  and  }  1/\gamma_g=\max_{k\in B_{\text{tight}}} \| \nabla X\left(k\right) \|
\end{equation*}
Finally, the three inter extreme points geodesics $G_{\text{deep}}=(G^{x},G^{y},G^{z})$ are computed using the Dijkstra algorithm \cite{Dijkstra} after each forward pass to update and refine our deep geodesics over training iterations.

\subsection{Pseudo-label propagation via CRF regularised loss}
In this section, we propose to propagate the pseudo-label information provided by the deep geodesics and the bounding box to the unlabelled voxels. Instead of using an external technique that requires a computationally expensive optimisation (e.g. CRFs, Random Walker) \cite{Rajchl:deepcut,Roth:dextr3d}, we propose to use a direct unsupervised regularised loss as employed in other works \cite{Tang:regularised,Dorent:scribbleda}. This regularisation term encourages spatial and intensity consistency and can be seen as a relaxation of the pairwise CRF term. It is defined as:
\begin{equation*}
    R(f_{\theta}\left(X\right)) \triangleq \frac{1}{|\Omega|}\sum_{k,l\in \Omega}  f_{\theta}(X)_k \exp \Big(-\frac{d(k,l)^{2}}{2 \sigma_{\alpha}^{2}}-\frac{(X_{k}-X_{l})^{2}}{2 \sigma_{\beta}^{2}}  \Big) (1-f_{\theta}(X)_l)
\end{equation*}
where $\sigma_{\alpha}$ and $\sigma_{\beta}$ respectively control the spatial and intensity consistency.

\subsection{Final model}
To summarise, deep geodesics are computed on the fly after each forward pass during training.
The voxels belonging to geodesics connecting extreme points ($G_{\text{deep}}$) and those outside of the relaxed bounding box ($\overline{B_{\text{relax}}}$) are used for supervision and a regularised loss ensures consistency:
\begin{equation}
    \theta^* = \argmin_{\theta} \sum_{i=1}^{n} \mathcal{L}(f_\theta(X_i), \overline{B_{\text{relax}}}\cup G_{\text{deep}}) + \lambda R(f_\theta(X_i))
\end{equation}
where $\mathcal{L}$ is a segmentation loss on the annotated voxels. We employed the sum of the Dice loss, the cross-entropy and the class-balanced focal loss.

\section{Experiments}
\subsubsection{Dataset and annotations.} We conducted experiments on a large publicly available dataset\footnote{\url{https://doi.org/10.7937/TCIA.9YTJ-5Q73 }} for Vestibular Schwannoma~(VS) segmentation using high-resolution T2 (hrT2) scans as input. VS is a non-cancerous tumour located on the main nerve connecting the brain and inner ear and is typically path-connected. This collection contains a labelled dataset of MR images collected on 242 patients with VS undergoing Gamma Knife stereotactic radiosurgery (GK SRS). hrT2 scans have an in-plane resolution of approximately $0.5 \times 0.5$mm, an in-plane matrix of $384 \times 384$ or $448 \times 448$, and slice thickness of $1.0-1.5$mm. The dataset was randomly split into 172 training, 20 validation and 46 test scans.
Manual segmentations were performed in consensus by the treating neurosurgeon and physicist. Manual extreme clicks were done by a biomedical engineer.

\subsubsection{Implementation details.} Our models were implemented in PyTorch using MONAI and TorchIO \cite{Perez:torchio}.
A 2.5D U-Net designed for VS segmentation was employed \cite{Shapey:jns,Wang:vsseg}. Mimicking training policies used in nnU-Net \cite{Isensee:nnunet}, 2.5D U-Nets were trained for 9000 iterations with a batch size of 6, i.e. 300 epochs of the full training set. Random patches of size $224\times224\times48$ were used during training. SGD with Nesterov momentum ($\mu=0.99$) and an initial learning rate of $0.01$ were used. The poly learning rate policy was employed to decrease the learning rate $(1-\frac{\lfloor \text{it}/30\rfloor}{300})^{0.9}$. The bounding box margin was set to $r=4$ voxels. For the regularised loss, we used a public implementation based on \cite{Joutard:lattice,Dorent:scribbleda}. $\sigma_{\alpha}$, $\sigma_{\beta}$ and $\lambda$ were respectively set to $15$, $0.05$ and $10^{-4}$. Similar results were obtained for $\sigma_{\alpha} \in \{ 5,15\}$ and $\sigma_{\beta}\in\{0.5,0.05\}$.
The model with the smallest validation loss is selected for evaluation. 
Our code and pre-trained models are publicly available\footnote{ \url{https://github.com/ReubenDo/InExtremIS/}}.

\subsubsection{Evaluation.} To assess the accuracy of each fully automated VS segmentation method, Dice Score (Dice) and $95$th percentile of the Hausdorff Distance (HD$95$) were used. Precision was also employed to determine whether an algorithm tends to over-segment. Wilcoxon signed rank tests ($p<0.01$) are performed. 

\begin{table*}[t!]
    \caption{Quantitative evaluation of different approaches for VS segmentation. Mean and variance are reported. Improvements in each stage of our ablation study are statistically significant $p<0.01$ as per a Wilcoxon test.}\label{tab:results}
    \begin{tabularx}{\textwidth}{X *{4}{c}}
        \toprule
        Approach & Dice ($\%$) & HD$95$ ($\text{mm}$) & Precision ($\%$)  \\
        \midrule
        Gradient Geodesic & 47.6 (15.5) & 37.7 (44.9) & 33.6 (15.0) \\
        Gradient+Euclidean Geodesic & 68.4 (14.9) & 65.3 (41.3) & 62.9 (23.1) \\
        Deep Geodesic (ours) & 78.1 (14.7) & 11.9 (25.5) & 88.3 (15.2) \\
        \textbf{Deep Geod. + Reg. Loss (InExtremIS)} & \textbf{81.9 (8.0)} & \textbf{3.7 (7.4)} & \textbf{92.9 (5.9)} \\
        \midrule
        Simulated Extreme Points (InExtremIS) & 83.7 (7.7) & 4.9 (14.9) & 90.0 (7.5) \\
        \midrule
        Fully Sup. - Equivalent Budget (2 hours) & 70.1 (14.2) & 8.9 (13.0) & 86.1 (13.1) \\
        Fully Sup. - Unlimited Budget (26 hours) & 87.3 (5.5) & 6.8 (19.9) & 84.7 (8.2) \\
        \midrule
        DeepCut \cite{Rajchl:deepcut} & 52.4 (30.2) & 12.9 (14.9) & 52.4 (29.6)\\
        Bounding Box Constraints \cite{Kervadec:boundingbox} &  56.0 (18.8) & 16.4 (17.5) & 49.7 (19.1) \\
        \bottomrule
    \end{tabularx}
\end{table*} 
\subsubsection{Ablation study.} To quantify the importance of each component of our framework, we conducted an ablation study. The results are shown in Table~\ref{tab:results}. We started with geodesics computed using only the image gradient magnitude, as in~\cite{Roth:dextr3d}.
Adding the Euclidean distance term significantly increased the Dice score by $+20.8$pp (percentage point). Adding the network probability term led to satisfying performance with a significant boost of $+9.7$pp in Dice score. In terms of training speed, the computation of the deep geodesics paths only took $0.36$sec per training iteration. Finally, using the CRF regularised loss allowed for more spatially consistent segmentations, as shown in Figure~\ref{qualitative}, increasing significantly the Dice by $+3.8$pp and reducing the HD$95$ by a factor 3.
These results demonstrate the role of each contribution and prove the effectiveness of our approach.

\subsubsection{Extreme points (37sec/scan) vs. full annotations (477sec/scan).} To compare the annotation time between weak and full supervision, a senior neurosurgeon segmented 5 random training scans using ITK-SNAP. On average it takes 477 seconds to fully annotate one hrT2 scan. In comparison, the full dataset was annotated with extreme points in 1h58min using ITK-SNAP, i.e. 37sec/scan. Full segmentation is thus $13\times$ more time-consuming than extreme points. Note that the reported annotation times cover the full annotation process, including the time required to open, save and adjust the image contrast ($\sim$15sec).

\subsubsection{Manual vs. simulated extreme points.} To study the robustness of our method to extreme points precision, we compared our approach using manual extreme points and extreme points obtained using the ground-truth segmentations. Since the simulated points are on the object boundaries, $r$ was set to $0$.
Table~\ref{tab:results} shows comparable results using both types of extreme points.

\begin{figure}[tb!]
\begin{center}
\includegraphics[width=\linewidth]{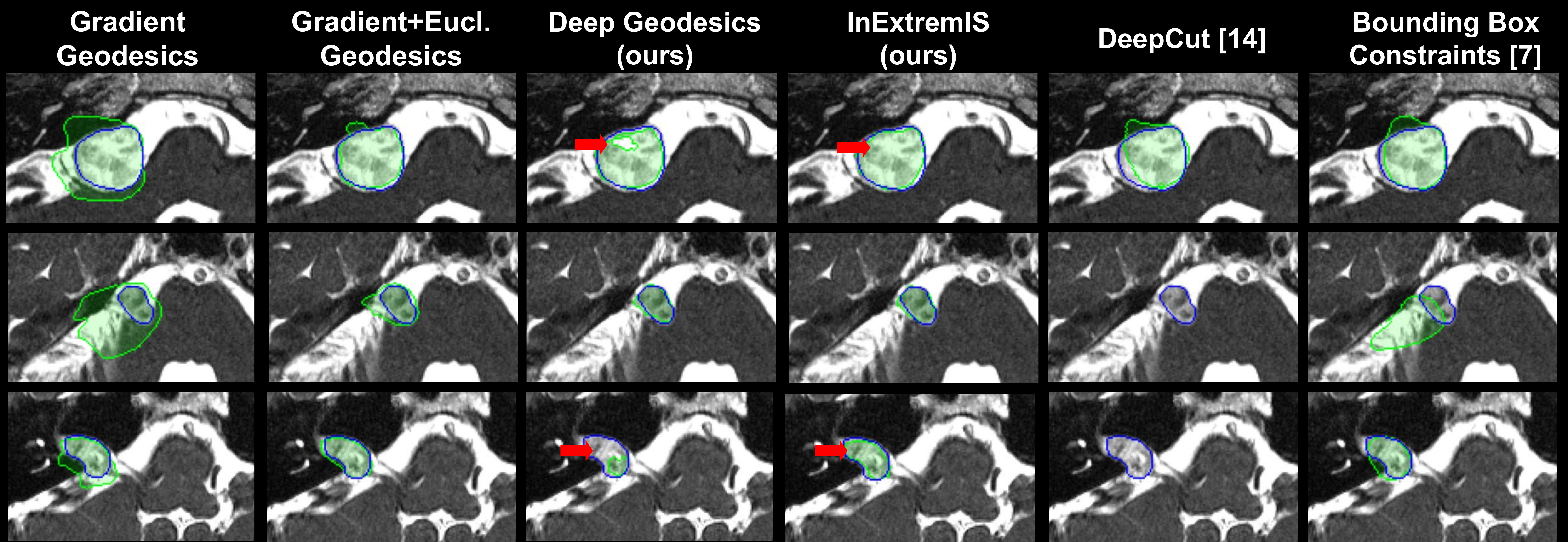}
\caption{Qualitative comparison of the different approaches. In green the predictions, in blue the manual segmentations. Red arrows show the effect of the regularised loss.}
\label{qualitative}
\end{center}
\end{figure}

\subsubsection{Segmentation accuracy per annotation budget.} We proposed to determine what is the best annotation approach given an annotation time budget. To do so, we trained a fully supervised approach using different time budgets. Given an annotation budget, the number of training and validation scans for full supervision was estimated by considering an annotation time of 477sec/scan. Results are detailed in Appendix. First, extreme point supervision outperformed  full supervision given a fixed time budget of 2 hours ($+11.8$pp Dice). Full-supervision even required a $5\times$ larger annotation budget than extreme point supervision to reach comparable performance. Finally, as expected, full supervision slightly outperformed weak supervision with an unlimited budget ($+5.4$pp Dice).

\subsubsection{Bounding boxes vs. extreme points.} We compared our approach to two weakly supervised approaches based on bounding boxes: DeepCut \cite{Rajchl:deepcut} and a recently proposed 2D approach using constraints \cite{Kervadec:boundingbox}. These approaches were implemented in 2D using the code from \cite{Kervadec:boundingbox}. A 2D U-Net was used to allow for a fair comparison. The dilated bounding boxes were used for training. Results in Table~\ref{tab:results} show that our approach significantly outperformed these techniques, demonstrating the benefits of our weakly supervised technique. Precision measurements and qualitative results in Figure~\ref{qualitative} highlight a key drawback of these approaches: They tend to over-segment when the bounding boxes are drawn in 3D and thus not tight enough for every single 2D slice. In contrast, our approach doesn't have any tightness prior and allow for large 2D box margins. Note that comparisons with semi-automated techniques requiring bounding boxes or extreme points at inference time are out-of-scope.

\section{Discussion and conclusion}
In this work, we presented a novel weakly supervised approach using only 6 extreme points as training annotation. We proposed to connect the extreme points using a new formulation of geodesics that integrates the network outputs. Associated with a CRF regularised loss, our approach outperformed other weakly supervised approaches and achieved competitive performance compared to a fully supervised approach with an unlimited annotation budget. A 5 times larger budget was required to obtain comparable performance using a fully supervised approach. Our approach was tested on manual interactions and not on simulated interactions as in \cite{Roth:dextr3d,Kervadec:boundingbox}. Our framework could typically be integrated in semi-automated pipelines to efficiently annotate large datasets with connected objects (e.g., VS, pancreas, liver, kidney). Future work will be conducted to identify which structures and shapes are suitable for our approach.

Although our approach has only been tested on a single class problem, the proposed technique is compatible with multi-class segmentation problems. Future work will especially investigate multi-class problems. We also plan to use our approach for weakly supervised domain adaptation problems.

\subsubsection{Acknowledgement}
This work was supported by the Engineering and Physical Sciences Research Council (EPSRC) [NS/A000049/1] and Wellcome Trust [203148/Z/16/Z]. TV is supported by a Medtronic / Royal Academy of Engineering Research Chair [RCSRF1819\textbackslash7\textbackslash34]. 

\bibliographystyle{splncs04}
\bibliography{paper1754}


\end{document}